\documentclass[wcp]{jmlr}

\usepackage{amsmath,amssymb,graphicx,url,doi}

\jmlrvolume{266}
\jmlryear{2025}
\jmlrworkshop{Conformal and Probabilistic Prediction with Applications}
\jmlrproceedings{PMLR}{Proceedings of Machine Learning Research}

\title{Stacked conformal prediction}
\author{\Name{Paulo C. Marques} F.
\Email{PauloCMF1@insper.edu.br} \\
\addr{Insper Institute of Education and Research, Rua Quatá 300, São Paulo 04546-042, Brazil}}

\editor{Khuong An Nguyen, Zhiyuan Luo, Tuwe Löfström, Lars Carlsson and Henrik Boström}

\begin{document}

\maketitle

\begin{abstract}
We consider a method for conformalizing a stacked ensemble of predictive models, showing that the potentially simple form of the meta-learner at the top of the stack enables a procedure with manageable computational cost that achieves approximate marginal validity without requiring the use of a separate calibration sample. Empirical results indicate that the method compares favorably to a standard inductive alternative.
\end{abstract}

\begin{keywords}
Supervised learning, prediction sets, conformal prediction, stacking, exchangeability.
\end{keywords}

\section{Introduction}\label{sec:intro}

In its most general transductive (full) form, conformal prediction \citep{vovk1999,vovk2005} creates a tension between desirable statistical properties and computational complexity: while it always guarantees, for exchangeable data, the finite-sample marginal coverage of the nonparametric, model-agnostic prediction sets produced by the method, it may, in certain settings, require a highly intensive computational process, due to the necessity of repeatedly retraining the underlying predictive model. Inductive (split) conformal prediction \citep{papadopoulos,lei} is an appealing and widely adopted variation that balances the computational cost through a trade-off: while preserving the statistical properties of the full conformal prediction procedure and requiring a single training process for the specified predictive model, it forces the available data to be split into training and calibration samples, which may result in neither building the most accurate predictive model, nor obtaining the most efficient confidence assessment of the model's generalization capacity. Several alternatives have been developed to keep the computational cost under control while making a more symmetric use of the available data, starting with cross-conformal prediction \citep{vovk2015}, which inspired the development of the jackknife+ \citep{barber} and the nested conformal prediction framework \citep{gupta}.

In this work, we develop a method for conformalizing a stacked ensemble of predictive models \citep{wolpert,breimanSR}. This construction serves a dual purpose. First, it leverages the usual predictive gains of model stacking. Second, it takes advantage of the fact that the meta-learner at the top of the stack can be a simpler model, enabling a low-cost conformalization procedure that eliminates the need for a separate calibration set while producing prediction intervals with approximate marginal validity.

We proceed in the following manner. In Section \ref{sec:symm}, in a regression setting with exchangeable data, we construct a stack of predictive models that becomes totally symmetric by including the future observable pair in the training sample used to fit the base-learners at the first stack level. For this symmetric stack, we prove that the assumed exchangeability of the training sample is transferred to the second level of the stack. Consequently, for an arbitrary meta-learner placed at the top of the stack, we have the exact marginal validity of a random conformal prediction set associated with the future observable pair. Section \ref{sec:feasible} makes the stack construction feasible by moving the future observable pair out of the training sample. This breaks the distributional symmetry of the stack, but, under a stability assumption for the base stack models, the previously stated marginal validity property still holds approximately. In Section \ref{sec:meta}, we specialize the meta-learner to multiple linear regression to take advantage of the fact that for this simple model a full conformalization procedure can be carried out at low computational cost through classical rank-one matrix updates. We apply the resulting stacked conformal prediction procedure to two datasets, comparing the stacked conformal prediction intervals to those produced by the inductive conformalized quantile regression method. In Section \ref{sec:code}, we present our concluding remarks and give pointers to open source software implementations of the developed method.

\section{Symmetric stack}\label{sec:symm}

In a supervised learning setting, we have random pairs ${\{(X_i,Y_i)\}_{i=1}^{n+1}}$, for which the features vector $X_i\in\mathbb{R}^d$ and the response variable $Y_i\in\mathbb{R}$. In this regression problem, we refer to $(X_{n+1},Y_{n+1})$ as the future observable pair. Let $\mathbb{S}_n$ denote the set of $n\times n$ permutation matrices, which are square matrices with exactly one entry equal to $1$ in each row and column, and all other entries equal to zero. Laying out our training sample --- which for now includes the future observable pair --- as the $(n+1)\times(d+1)$ random matrix
\[
  T = \begin{pmatrix}
  X_{1,1} & X_{1,2} & \cdots & X_{1,d} & Y_{1} \\
  X_{2,1} & X_{2,2} & \cdots & X_{2,d} & Y_{2} \\
  \vdots  & \vdots  & \ddots & \vdots  & \vdots \\
  X_{n,1} & X_{n,2} & \cdots & X_{n,d} & Y_{n} \\
  X_{n+1,1} & X_{n+1,2} & \cdots & X_{n+1,d} & Y_{n+1}
  \end{pmatrix},
\]
we make the assumption that the data are exchangeable, meaning that $T$ and $\Pi T$ have the same distribution, for every permutation matrix $\Pi\in\mathbb{S}_{n+1}$.

We randomly divide the training sample into $K\geq 2$ folds of size $t=(n+1)/K$, assuming that $n+1$ is divisible by $K$, by means of an $(n+1)\times (n+1)$ random permutation matrix $Q$, which we refer to as the folding scheme matrix. We suppose that $Q$ is uniformly distributed, so that $\Pr(Q=q)=1/(n+1)!$, for every $q\in\mathbb{S}_{n+1}$, and we assume that this folding scheme matrix $Q$ and the training data $T$ are independent. We use the block notation
\[
  Q = \begin{pmatrix} Q_1 \\ Q_2 \\ \vdots \\ Q_K \end{pmatrix},
\]
in which the realizations of the matrix blocks $Q_k$ are in $\mathbb{R}^{t\times(n+1)}$, for $k=1,\dots,K$. For the folding scheme defined by the matrix $Q$, let $\phi$ be a fold indicator function, defined by $\phi(i)=\sum_{k=1}^K k \cdot I(\sum_{\ell=1}^t (Q_k)_{\ell,i}=1)$, meaning that $\phi(i)$ determines the fold number $k$ to which the $i$-th training sample unit has been assigned, for $i=1,\dots,n+1$. Moreover, using the notation introduced for the $Q$ blocks, we define, for $k=1,\dots,K$, the $k$-th fold exclusion matrix
\[
  Q_{\setminus k} = \begin{pmatrix} Q_1 \\ \vdots \\ Q_{k-1} \\ Q_{k+1} \\ \vdots \\  Q_K \end{pmatrix},
\]
whose realizations are in $\mathbb{R}^{(n+1-t)\times(n+1)}$.

We define a symmetric stacked ensemble of predictive models as follows. The stack is built from $M\geq 1$ base learning methods. For each training sample unit, the goal is to use its features vector to predict the corresponding response variable using models trained only on training data folds that do not include the training sample unit under consideration. Formally, for $m=1,\dots,M$, we have prediction functions
\[
  \hat\mu_m:\mathbb{R}^{(n+1-t)\times(d+1)}\times\mathbb{R}^d\to\mathbb{R},
\]
and we assume that each learning method treats its training data $S\in\mathbb{R}^{(n+1-t)\times(d+1)}$ symmetrically, so that
\begin{equation}\label{eq:muhat}
  \hat\mu_m(S,x)=\hat\mu_m(\Pi S,x),
  \tag{$\star$}
\end{equation}
for every permutation matrix $\Pi\in\mathbb{S}_{n+1-t}$ and each $x\in\mathbb{R}^d$. When the learning method involves some form of randomization --- such as the bootstrap process in Random Forests \citep{breimanRF}, or the stochastic gradient descent optimization in Deep Neural Networks \citep{bishop} --- we assume, without loss of generality, that the seed of the underlying pseudo-random number generator is set using a symmetric hash function of the training data $S$. The stack base-learners make predictions
\[
  Z_i = (Z_{i,1},\dots, Z_{i,M}) \in \mathbb{R}^M,
\]
in which $Z_{i,m} = \hat\mu_m\!\left(Q_{\setminus \phi(i)}T,X_i\right)$, for $i=1,\dots,n+1$. This totally symmetric definition of the stack implies that the assumed exchangeability of the training sample $T$ is transferred to the next stack level, as stated by the following result, proved in the Appendix.

\begin{proposition}\label{prop:exch}
The second-level random pairs $(Z_1,Y_1),\dots,(Z_{n+1},Y_{n+1})$ are exchangeable for a symmetric stack.
\end{proposition}

The following result, proved in the Appendix, formalizes the idea that, for a generic meta-learner placed at the top of the symmetric stack, Proposition \ref{prop:exch} and a standard conformal argument allow us to construct a random prediction set for the future observable pair with exact marginal validity. We denote the ceiling of $u\in\mathbb{R}$ by $\lceil u\rceil=\min\{z\in\mathbb{Z}:u\leq z\}$.

\begin{proposition}\label{prop:mvp}
Let $\hat\psi_n^{(y)}$ be a meta-learner trained from $\{(Z_1,Y_1),\dots,(Z_n,Y_n),(Z_{n+1},y)\}$, for $y\in\mathbb{R}$, and suppose the order of the $n+1$ pairs in this sample is irrelevant for the construction of $\hat\psi_n^{(y)}$. For a conformity function $\rho:\mathbb{R}\times\mathbb{R}\to\mathbb{R}$, define the conformity scores $R_i^{(y)} = \rho(Y_i,\hat\psi^{(y)}_n(Z_i))$, for $i=1,\dots,n+1$, and let $R^{(y)}_{(1)}\leq R^{(y)}_{(2)}\leq\dots\leq R^{(y)}_{(n)}$ denote the ordered conformity scores among $\{R^{(y)}_1,R^{(y)}_2\dots,R^{(y)}_n\}$. Choosing a nominal miscoverage level $0<\alpha<1$ such that $\lceil(1-\alpha)(n+1)\rceil\leq n$, we have that
\[ \Pr\!\left( Y_{n+1} \in C^{(\alpha)}_{n+1} \right) \geq 1 - \alpha, \]
in which we defined the random prediction set $C^{(\alpha)}_{n+1} = \left\{ y\in\mathbb{R} : R_{n+1}^{(y)} \leq R_{(\lceil(1-\alpha)(n+1)\rceil)}^{(y)} \right\}$.
\end{proposition}

\section{Feasible stack}\label{sec:feasible}

The symmetric stack described in Section \ref{sec:symm} is an oracle construct that cannot be implemented in practice, since at training time we do not know the observed value of the future response $Y_{n+1}$. A \textit{feasible stack} is attained by removing the future observable pair $(X_{n+1},Y_{n+1})$ from the training sample. In doing so, we break the distributional symmetry of the stack, but if the predictions made by the stack base-learners stay stable after this single sample unit removal, we can argue that the marginal validity property stated in Proposition \ref{prop:mvp} still holds approximately, as will be corroborated by the empirical results in Section \ref{sec:meta}.

Suppose that in this feasible stack the first level prediction $Z_{n+1}$ is now made by base-learners trained on the whole training sample. Following the notations introduced in Proposition \ref{prop:mvp} for the symmetric stack, let $\tilde R_{n+1}^{(Y_{n+1})}$ and $\tilde R_{(\lceil(1-\alpha)(n+1)\rceil)}^{(Y_{n+1})}$ be the corresponding random conformity scores pertaining to the feasible stack.

The following approximation result, proved in the Appendix, shows how the marginal validity of the stacked conformal prediction procedure is impacted by the broken distributional symmetry of the feasible stack: if the base-learners are stable in the sense that we can control the probability of the corresponding conformity scores on both the symmetric and the feasible stack differing too much, the marginal validity property holds approximately for the feasible stack.  

\begin{proposition}\label{prop:approx}
Given $\epsilon>0$, if there is a $\delta=\delta(\epsilon)>0$ such that
\[
  \Pr\!\left(\max\left\{\left|\tilde R_{n+1}^{(Y_{n+1})} - R_{n+1}^{(Y_{n+1})}\right|,\left|\tilde R_{(\lceil(1-\alpha)(n+1)\rceil)}^{(Y_{n+1})} - R_{(\lceil(1-\alpha)(n+1)\rceil)}^{(Y_{n+1})}\right|\right\}<\epsilon/2\right)\geq 1-\delta, 
\]
\noindent then
\[
  \Pr\!\left( Y_{n+1} \in \tilde C^{(\alpha)}_{n+1} \right) \geq 1 - \alpha - \delta - h(\epsilon),
\]
\noindent \\
in which
\[
  \tilde C^{(\alpha)}_{n+1} = \left\{ y\in\mathbb{R} : \tilde R_{n+1}^{(y)} \leq \tilde R_{(\lceil(1-\alpha)(n+1)\rceil)}^{(y)} \right\}
\]
\noindent \\
and 
\[
  h(\epsilon) = \Pr\!\left(R_{(\lceil(1-\alpha)(n+1)\rceil)}^{(Y_{n+1})} - \epsilon < R_{n+1}^{(Y_{n+1})} \leq R_{(\lceil(1-\alpha)(n+1)\rceil)}^{(Y_{n+1})}\right).
\]
\end{proposition}

\section{Specializing the meta-learner}\label{sec:meta}

\begin{algorithm2e}[t!]
\caption{Full conformal prediction for multiple linear regression}\label{algo:fullcp}
\SetArgSty{textbf}
\SetAlgoVlined
\DontPrintSemicolon
\LinesNumbered
\KwIn{Matrix $Z\in\mathbb{R}^{n\times M}$ of second-level stack predictions for the training sample. Vector $y\in\mathbb{R}^{n\times 1}$ of training sample responses. Matrix $Z_0\in\mathbb{R}^{m\times M}$ of second-level stack predictions for the size $m$ test sample. Nominal miscoverage level $0<\alpha<1$. Tolerance $\epsilon>0$ and multiple $u>0$ for searching interval limits.}
\KwOut{Matrix $C\in\mathbb{R}^{m\times 2}$ of lower and upper prediction interval limits for the test sample.}
\BlankLine
$\texttt{sd}\gets\text{standard deviation of $y$}$\;
$A\gets(Z^\top Z)^{-1}$\;
$\hat\beta\gets A Z^\top y$\;
Initialize $C=(c_{ij})\in\mathbb{R}^{m\times 2}$ with zeroes\;
\For{$i\leftarrow 1$ \KwTo $m$}{
  $z_0\gets\text{transpose of $i$-th line of $Z_0$}$\;
  $B\gets A - (Az_0z_0^\top A)/(1+z_0^\top Az_0)$\; \label{algo:fullcp:sm}
  $\texttt{inf}\gets z_0^\top\hat\beta$\;
  $\texttt{lower}\gets \texttt{inf} - u\times \texttt{sd}$\;
  \While{$\texttt{inf}-\texttt{lower}>\epsilon$}{
    $y_0\gets(\texttt{lower}+\texttt{inf})/2$\;
    Get scores $r_0$ and $\hat r$ inputing $(z_0, y_0, Z, y, \alpha, A, \hat\beta, B)$ to Algorithm \ref{algo:scores}\;
    \If{$r_0\leq\hat r$}{
      $\texttt{inf}\gets y_0$\;
    } \Else{
      $\texttt{lower}\gets y_0$\;
    }
  }
  $c_{i,1}\gets\texttt{inf}$\;
  $\texttt{sup}\gets z_0^\top\hat\beta$\;
  $\texttt{upper}\gets \texttt{sup} + u\times \texttt{sd}$\;
  \While{$\texttt{upper}-\texttt{sup}>\epsilon$}{
    $y_0\gets(\texttt{sup}+\texttt{upper})/2$\;
    Get scores $r_0$ and $\hat r$ inputing $(z_0, y_0, Z, y, \alpha, A, \hat\beta, B)$ to Algorithm \ref{algo:scores}\;
    \If{$r_0\leq\hat r$}{
      $\texttt{sup}\gets y_0$\;
    } \Else{
      $\texttt{upper}\gets y_0$\;
    }
  }
  $c_{i,2}\gets\texttt{sup}$
}
\Return{$C$}
\end{algorithm2e}

The feasible stack described in Section \ref{sec:feasible} is useful only if we can make the computational cost involved in the conformalization of the meta-learner manageable. It has long been known \citep{vovk2005} that full conformalization of a multiple linear regression model can be carried out efficiently through rank-one matrix updates made with the help of the Sherman-Morrison formula \citep{sherman}, which states that, if the matrix $A\in\mathbb{R}^{n\times n}$ is invertible and $u,v\in\mathbb{R}^{n\times 1}$, then $A+uv^\top$ is invertible if and only if $1+v^\top A^{-1}u\neq0$, in which case its inverse can be computed efficiently from $A^{-1}$ as
\[
  (A + uv^\top)^{-1} = A^{-1} - \frac{A^{-1}uv^\top A^{-1}}{1 + v^\top A^{-1}u}.
\]

Algorithms \ref{algo:fullcp} and \ref{algo:scores} formally describe the efficient conformalization of a multiple linear regression meta-learner based on this classical formula (see line \ref{algo:fullcp:sm} of Algorithm \ref{algo:fullcp}). It is worth mentioning that the residuals determination in Algorithm \ref{algo:scores} do not involve the computation of additional inverses. The use of these residuals to construct the conformity scores denominators is crucial for obtaining prediction intervals with more adaptive width.

\begin{algorithm2e}[t!]
\caption{Conformity scores computation}\label{algo:scores}
\SetAlgoVlined
\DontPrintSemicolon
\LinesNumbered
\KwIn{Values of $z_0\in\mathbb{R}^{M\times 1}$, $y_0\in\mathbb{R}$, $Z\in\mathbb{R}^{n\times M}$, $y\in\mathbb{R}^{n\times 1}$, $0<\alpha<1$, $A\in\mathbb{R}^{M\times M}$, $\hat\beta\in\mathbb{R}^{M\times 1}$, and $B\in\mathbb{R}^{M\times M}$ from Algorithm \ref{algo:fullcp}.}
\KwOut{Conformity scores $r_0$ and $\hat r$.}
\BlankLine
$\hat\beta_0 \gets \hat\beta + (y_0 - z_0^\top \hat\beta)\,B z_0$\; % \tcp*{$\hat\beta_0\in\mathbb{R}^{M\times 1}$} % remove \;
$\hat y \gets Z\hat\beta_0$\;
$\hat y_0 \gets z_0^\top \hat\beta_0$\;
Initialize $y^{(\text{res})}\in\mathbb{R}^{n\times 1}$ with zeroes\;
\For{$i\leftarrow 1$ \KwTo $n$}{
  $y^{(\text{res})}_i \gets |y_i - \hat y_i|$\;
}
$y^{(\text{res})}_0 \gets |y_0-\hat y_0|$\;
$\hat\beta^{(\text{res})} \gets A Z^\top y^{(\text{res})}$\;
$\hat\beta^{(\text{res})}_0 \gets \hat\beta^{(\text{res})} + (y^{(\text{res})}_0 - z_0^\top \hat\beta^{(\text{res})})\,B z_0$\;
$\hat\delta \gets Z \hat\beta^{(\text{res})}_0$\;
$\hat\delta_0 \gets z_0^\top \hat\beta^{(\text{res})}_0$\;
Initialize $r\in\mathbb{R}^{n\times 1}$ with zeroes\;
\For{$i\leftarrow 1$ \KwTo $n$}{
  $r_i \gets y^{(\text{res})}_i / (1 + \hat\delta_i)$\;
}
$\hat r \gets r_{(\lceil(1-\alpha)(n+1)\rceil)}$\;
$r_0 \gets y^{(\text{res})}_0 / (1 + \hat\delta_0)$\; 
\Return{$(r_0,\hat r)$}
\end{algorithm2e}

We implemented the stacked conformal prediction procedure for two examples. The California housing dataset \citep{pace} comprises information on 20,640 census tracts in California, drawn from the 1990 U.S. Census. It includes eight predictors and the response variable is the median house value in US dollars for each tract. The Ames housing dataset \citep{decock} contains information on 2,930 houses sold in Ames, Iowa from 2006 to 2010. For the Ames dataset, we have eighty predictors and the response variable is the sale price of the house in US dollars.

\begin{figure}[t!]
\centering
\includegraphics[width=15.25cm]{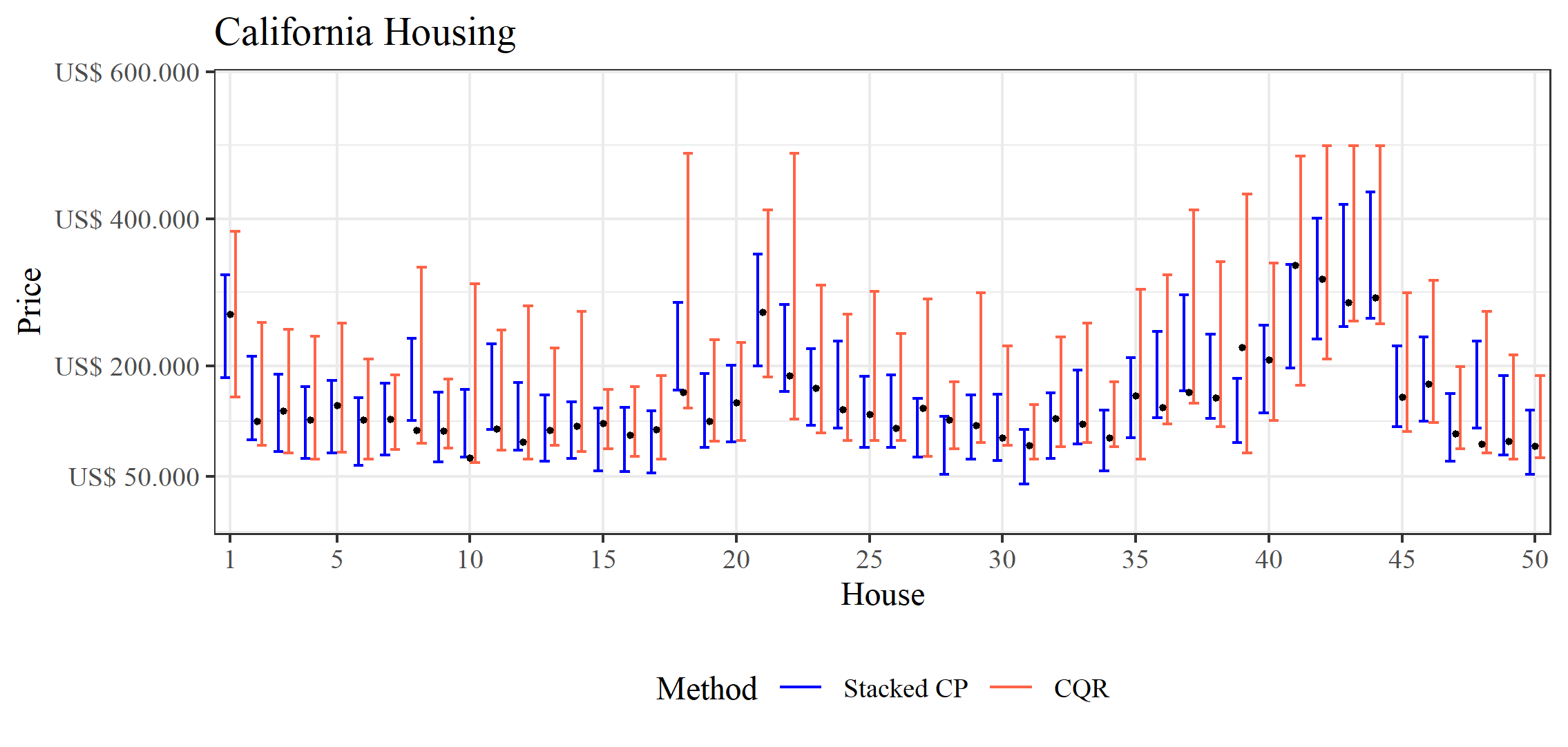}
\caption{Prediction intervals for fifty test sample units in the California Housing dataset, using a 90\% nominal coverage level ($\alpha=0.1$).}
\label{fig:california}
\end{figure}

\begin{figure}[t!]
\centering
\includegraphics[width=15.25cm]{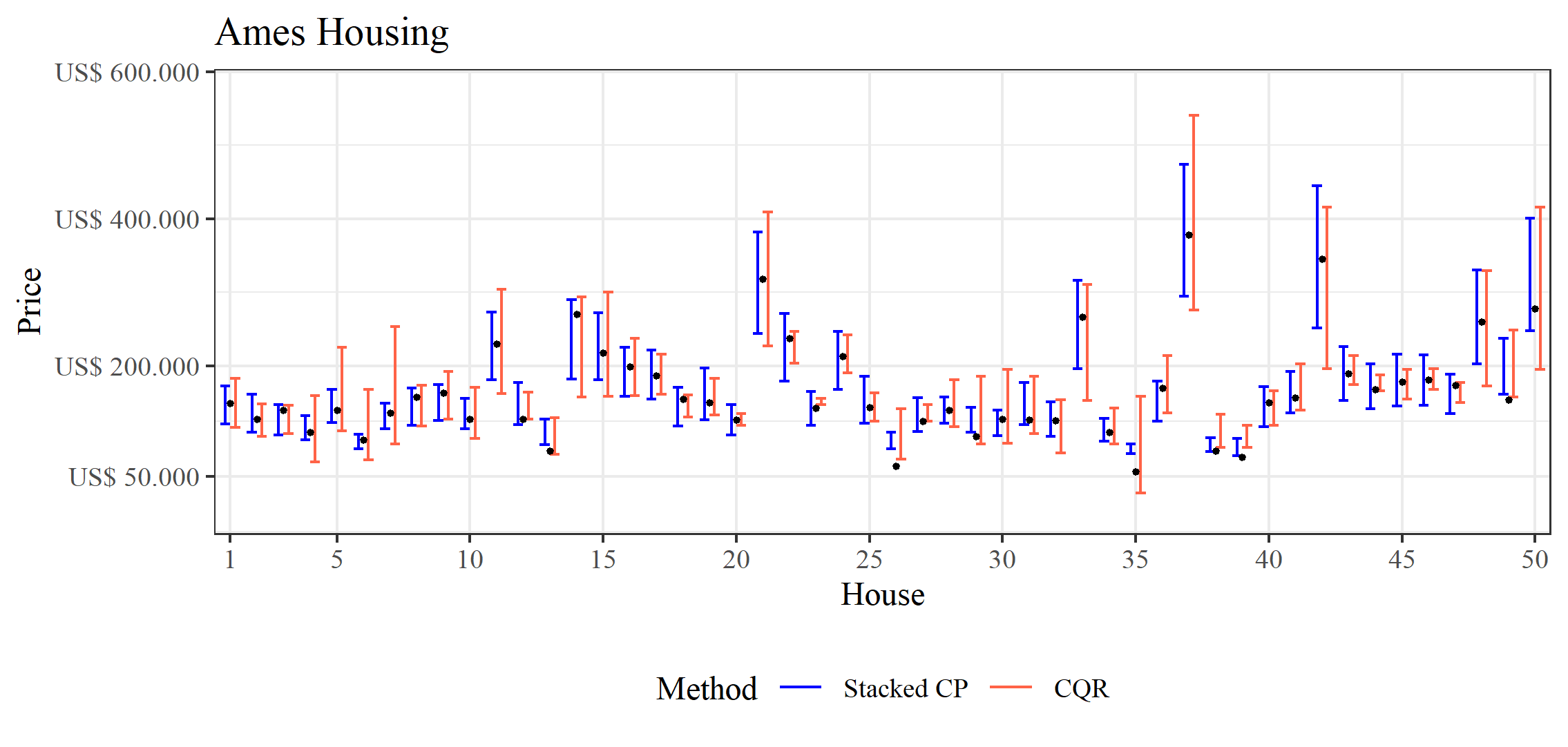}
\caption{Prediction intervals for fifty test sample units in the Ames Housing dataset, using a 90\% nominal coverage level ($\alpha=0.1$).}
\label{fig:ames}
\end{figure}

We used Random Forests \citep{breimanRF} and CatBoost \citep{catboost} as the stack base-learners. For both datasets, we used 70\% of the available data as our training sample, and generated prediction sets for the remaining 30\% test sample units. For comparison, we used conformalized quantile regression (CQR) \citep{romano} as an inductive alternative. The Quantile Random Forest algorithm \citep{meinshausen} was used in the implementation of CQR. For CQR, the calibration samples sizes were chosen to be 2,000 and 500, for the California and Ames datasets, respectively, following the calibration sample size discussion in \cite{coverage}. For each dataset, Figures \ref{fig:california} and \ref{fig:ames} show the prediction intervals for fifty randomly chosen test sample units, using a 90\% nominal coverage level ($\alpha=0.1$). Table \ref{tab:comparison} summarizes the results for different nominal coverage levels, showing that the stacked conformal prediction intervals exhibited appropriate empirical coverage, with shorter median interval width, when compared to the CQR results.

{\def\arraystretch{1.5}
\begin{table}[t!]
\centering
\small
\caption{Empirical coverage and prediction interval width summaries (US dollars) for different nominal miscoverage levels $\alpha$.}
\label{tab:comparison}
\begin{tabular}{llccccc}
Dataset    & Method     & $1-\alpha$ & Empirical coverage & 1st quartile & Median           & 3rd quartile \\ \hline
California & Stacked CP & 80\%       & 80.9\%             & 69,226       & \textbf{84,995}  & 105,696      \\
California & CQR        & 80\%       & 80.1\%             & 67,485       & 105,580          & 162,900      \\ \hline
California & Stacked CP & 85\%       & 85.5\%             & 80,864       & \textbf{99,283}  & 123,470      \\
California & CQR        & 85\%       & 85.7\%             & 81,111       & 125,310          & 192,075      \\ \hline
California & Stacked CP & 90\%       & 89.9\%             & 96,927       & \textbf{119,003} & 147,988      \\
California & CQR        & 90\%       & 90.8\%             & 104,433      & 156,600          & 231,500      \\ \hline
Ames       & Stacked CP & 80\%       & 81.9\%             & 27,665       & \textbf{39,511}  & 58,467       \\
Ames       & CQR        & 80\%       & 81.2\%             & 29,796       & 48,355           & 79,813       \\ \hline
Ames       & Stacked CP & 85\%       & 86.7\%             & 32,645       & \textbf{46,624}  & 69,001       \\
Ames       & CQR        & 85\%       & 85.9\%             & 34,401       & 55,210           & 91,590       \\ \hline
Ames       & Stacked CP & 90\%       & 91.1\%             & 41,652       & \textbf{59,478}  & 88,088       \\
Ames       & CQR        & 90\%       & 90.2\%             & 44,530       & 68,960           & 108,839      \\ \hline
\end{tabular}
\end{table}}

\section{Concluding remarks and open source implementations}\label{sec:code}

We have presented the stacked conformal prediction procedure in a regression setting, but its general structure readily extends to other tasks --- classification being a prime example. The key requirement in any such extension is the choice of a meta-learner that admits full conformalization at manageable computational cost. In classification problems, a natural meta-learner candidate would be the $k$-nearest neighbors classifier, equipped with a distance metric tailored to aggregate the class-probability outputs produced by the stack base-learners. \texttt{R} \citep{R} and Python \citep{python} code, including C\texttt{++} \citep{cpp} extensions, for full reproduction of the examples in the paper are available at:
\begin{center}
\texttt{https://github.com/paulocmarquesf/stacked\_cp} \quad
\end{center}

\acks{Paulo C. Marques F. receives support from FAPESP (Fundação de Amparo à Pesquisa do Estado de São Paulo) through project 2023/02538-0.}

\bibliography{bibliography.bib}

\newpage

\appendix

\section*{Appendix: Proofs}\label{appendix}

\setcounter{theorem}{0}

\renewcommand{\proofname}{Proof of Proposition \ref{prop:exch}}

\begin{proof}
Laying out the second-level pairs as the $(n+1)\times(M+1)$ matrix

{\footnotesize
\addtolength\jot{8pt}
\begin{align*}
  U = g(T,Q) &= \begin{pmatrix}
    Z_1 & Y_1 \\
    Z_2 & Y_2 \\
    \vdots & \vdots \\
    Z_n & Y_n \\
    Z_{n+1} & Y_{n+1}
  \end{pmatrix}
  =
  \begin{pmatrix}
  Z_{1,1} & Z_{1,2} & \cdots & Z_{1,M} & Y_{1} \\
  Z_{2,1} & Z_{2,2} & \cdots & Z_{2,M} & Y_{2} \\
  \vdots  & \vdots  & \ddots & \vdots  & \vdots \\
  Z_{n,1} & Z_{n,2} & \cdots & Z_{n,M} & Y_{n} \\
  Z_{n+1,1} & Z_{n+1,2} & \cdots & Z_{n+1,M} & Y_{n+1}
  \end{pmatrix} \\
  &= \begin{pmatrix}
  \hat\mu_1\!\left(Q_{\setminus \phi(1)}T,X_1\right) & \hat\mu_2\!\left(Q_{\setminus \phi(1)}T,X_1\right) & \cdots & \hat\mu_M\!\left(Q_{\setminus \phi(1)}T,X_1\right) & Y_{1} \\
  \hat\mu_1\!\left(Q_{\setminus \phi(2)}T,X_2\right) & \hat\mu_2\!\left(Q_{\setminus \phi(2)}T,X_2\right) & \cdots & \hat\mu_M\!\left(Q_{\setminus \phi(2)}T,X_2\right) & Y_{2} \\
  \vdots  & \vdots  & \ddots & \vdots  & \vdots \\
    \hat\mu_1\!\left(Q_{\setminus \phi(n)}T,X_n\right) & \hat\mu_2\!\left(Q_{\setminus \phi(n)}T,X_n\right) & \cdots & \hat\mu_M\!\left(Q_{\setminus \phi(n)}T,X_n\right) & Y_{n} \\
  \hat\mu_1\!\left(Q_{\setminus \phi(n+1)}T,X_{n+1}\right) & \hat\mu_2\!\left(Q_{\setminus \phi(n+1)}T,X_{n+1}\right) & \cdots & \hat\mu_M\!\left(Q_{\setminus \phi(n+1)}T,X_{n+1}\right) & Y_{n+1}
  \end{pmatrix},
\end{align*}}

\medskip

\noindent choose any permutation matrix $\Pi\in\mathbb{S}_{n+1}$, and consider the permutation of the $U$ rows defined by $\Pi U=\Pi g(T,Q)$. This permuted matrix $\Pi g(T,Q)$ can be obtained in a second way. Consider the folding scheme matrix defined by $Q\Pi^{-1}$. Because $(Q\Pi^{-1})(\Pi T)=QT$, this folding scheme matrix $Q\Pi^{-1}$ has the special property that it assigns the training sample units in the permuted training sample $\Pi T$ to the exact same folds that they had been assigned by the folding scheme matrix $Q$ applied to the original training sample $T$. Therefore, the assumed training data symmetry (\ref{eq:muhat}) of the base learners $\hat\mu_m$ in the definition of the symmetric stack yields the property: ${\Pi g(T, Q) = g(\Pi T, Q\Pi^{-1})}$. Since the class of permutation matrices is closed under multiplication and taking inverses, we have, for every $q\in\mathbb{S}_{n+1}$, that $\Pr(Q\Pi^{-1} = q) = \Pr(Q = q\Pi) = 1/(n+1)!$, implying that $Q$ and $Q\Pi^{-1}$ have the same distribution. Because we assumed that the training sample $T$ is exchangeable, and that $T$ and the folding scheme matrix $Q$ are independent, we come to the conclusion that $(T,Q)$ and $(\Pi T, Q\Pi^{-1})$ have the same distribution. Hence, $U=g(T,Q)$ and $\Pi U=\Pi g(T,Q) = g(\Pi T, Q\Pi^{-1})$ have the same distribution, as desired.
\end{proof}

The following result is used in the proof of Proposition \ref{prop:mvp}.

\begin{lemma}\label{lmm:order}
Let $U_1,U_2,\dots,U_n,U_{n+1}$ be exchangeable random variables, and denote the order statistics of the subset $\{U_1,U_2,\dots,U_n\}$ by $V_{(1)}\leq V_{(2)}\leq\dots\leq V_{(n)}$. For each ${k=1,\dots,n}$, it follows that $P(U_{n+1}\leq V_{(k)})\geq k/(n+1)$.
\end{lemma}

\renewcommand{\proofname}{Proof}

\begin{proof}
Denote the order statistics of the full set $\{U_1,U_2,\dots,U_n,U_{n+1}\}$ by
\[
  U_{(1)}\leq U_{(2)}\leq\dots\leq U_{(n)}\leq U_{(n+1)}.
\]
Considering that we may have ties, it follows from the definition of $U_{(k)}$ that at least $k$ of the $U_i$'s are less than or equal to $U_{(k)}$, for $k=1,\dots,n+1$. Hence, ${\sum_{i=1}^{n+1} I_{\{U_i\leq U_{(k)}\}}\geq k}$, almost surely. Taking expectations and observing that, by exchangeability, the probability ${P(U_i\leq U_{(k)})}$ is the same, for ${i=1,\dots,n+1}$, we have that ${P(U_{n+1}\leq U_{(k)})\geq k/(n+1)}$. Furthermore, for $k=1,\dots,n$, notice that $U_{n+1}>V_{(k)}$ if and only if $U_{n+1}>U_{(k)}$, because $U_{n+1}$ cannot be strictly larger than itself. Hence, ${P(U_{n+1}\leq V_{(k)})=P(U_{n+1}\leq U_{(k)})}$, for $k=1,\dots,n$, and the result follows.
\end{proof}

\renewcommand{\proofname}{Proof of Proposition \ref{prop:mvp}}

\begin{proof}
Taking $y=Y_{n+1}$, the definitions in the proposition statement and the result in Proposition \ref{prop:exch} imply that the random conformity scores
\[ R_1^{(Y_{n+1})}, R_2^{(Y_{n+1})}, \dots, R_n^{(Y_{n+1})}, R_{n+1}^{(Y_{n+1})} \]
are exchangeable. Using Lemma \ref{lmm:order} above, we have that
\[ \Pr\!\left(R^{(Y_{n+1})}_{n+1}\leq R^{(Y_{n+1})}_{(k)}\right) \geq \frac{k}{n+1}, \]
for $k=1,\dots,n$. Choosing $k=\lceil(1-\alpha)(n+1)\rceil\geq(1-\alpha)(n+1)$, the result follows, since the definition of the random prediction set $C_{n+1}^{(\alpha)}$ entails that $Y_{n+1}\in C_{n+1}^{(\alpha)}$ if and only if $R^{(Y_{n+1})}_{n+1}\leq R^{(Y_{n+1})}_{(\lceil(1-\alpha)(n+1)\rceil)}$.
\end{proof}

\renewcommand{\proofname}{Proof of Proposition \ref{prop:approx}}

\begin{proof}
Introduce the notations
\[
  r=R_{(\lceil(1-\alpha)(n+1)\rceil)}^{(Y_{n+1})} \quad \text{and} \quad \tilde r=\tilde R_{(\lceil(1-\alpha)(n+1)\rceil)}^{(Y_{n+1})},
\]
and define the event
\[
  B_\epsilon=\left\{ \max\left\{\left|\tilde R_{n+1}^{(Y_{n+1})} - R_{n+1}^{(Y_{n+1})}\right|,\left|\tilde r - r\right|\right\}<\epsilon/2 \right\}.
\]
If the statement in $B_\epsilon$ is true and $R_{n+1}^{(Y_{n+1})}\leq r - \epsilon$, then $\tilde R_{n+1}^{(Y_{n+1})} < R_{n+1}^{(Y_{n+1})} + \epsilon/2 \leq r - \epsilon/2 < \tilde r$.

\noindent Therefore, using the lower bound in Proposition \ref{prop:mvp}, the result follows:
\begin{align*}
  \Pr\!\left(Y_{n+1}\in \tilde C_{n+1}^{(\alpha)}\right) &= \Pr\!\left(\tilde R_{n+1}^{(Y_{n+1})}\leq \tilde r\right) \\
  &\geq \Pr\!\left(\left\{\tilde R_{n+1}^{(Y_{n+1})}< \tilde r\right\}\cap B_\epsilon\right) \\
  &\geq \Pr\!\left(\left\{R_{n+1}^{(Y_{n+1})}\leq r-\epsilon\right\}\cap B_\epsilon\right) \\
  &= \Pr\!\left(R_{n+1}^{(Y_{n+1})}\leq r-\epsilon\right) - \Pr\!\left(\left\{R_{n+1}^{(Y_{n+1})}\leq r-\epsilon\right\}\cap B^c_\epsilon\right) \\
  &\geq \Pr\!\left(R_{n+1}^{(Y_{n+1})}\leq r-\epsilon\right) - \Pr\!\left(B^c_\epsilon\right) \\
  &\geq \Pr\!\left(R_{n+1}^{(Y_{n+1})}\leq r-\epsilon\right) - \delta \\
  &= \Pr\!\left(R_{n+1}^{(Y_{n+1})}\leq r\right) - \Pr\!\left(r-\epsilon<R_{n+1}^{(Y_{n+1})}\leq r\right) - \delta \\
  &= \Pr\!\left(Y_{n+1}\in C_{n+1}^{(\alpha)}\right) - \Pr\!\left(r-\epsilon<R_{n+1}^{(Y_{n+1})}\leq r\right) - \delta \\
  &\geq 1 - \alpha - \delta - \Pr\!\left(r-\epsilon<R_{n+1}^{(Y_{n+1})}\leq r\right) \\
  &= 1 - \alpha - \delta - h(\epsilon). \\
\end{align*}
\end{proof}

\renewcommand{\proofname}{Proof.}

\end{document}